\documentclass{article}

\usepackage[preprint]{neurips_2020}

\usepackage[hyphens,spaces,obeyspaces]{url}
\usepackage{graphicx}
\usepackage{caption}
\usepackage{xcolor}
\usepackage{colortbl}
\usepackage{subcaption}
\usepackage{float}
\usepackage{verbatimbox}
\usepackage{algorithm}
\usepackage[margin=0.3cm]{caption}
\usepackage[noend]{algpseudocode}

\algrenewcommand\algorithmicforall{\textbf{foreach}}
\algrenewcommand\algorithmicindent{.8em}
\usepackage{booktabs}
\usepackage{multirow}
\usepackage{natbib}
\usepackage{listings}
\usepackage[linguistics]{forest}
\usepackage{dsfont}
\usepackage{caption}
\usepackage{amsmath}
\usepackage{hyperref}
\hypersetup{
    colorlinks = true,
    linkcolor = black,
    citecolor = blue,
    linkbordercolor = {white},
    urlcolor=blue,
}
\usepackage[utf8]{inputenc} % allow utf-8 input
\usepackage[T1]{fontenc}    % use 8-bit T1 fonts
\usepackage{hyperref}       % hyperlinks
\usepackage{url}            % simple URL typesetting
\usepackage{booktabs}       % professional-quality tables
\usepackage{amsfonts}       % blackboard math symbols
\usepackage{nicefrac}       % compact symbols for 1/2, etc.
\usepackage{microtype}      % microtypography
\RequirePackage[textsize=tiny,backgroundcolor=orange!20,linecolor=red!50]{todonotes}%

\title{Gaussian Process Molecular Property Prediction with FlowMO}

\author{%
  Henry B. Moss\thanks{Equal contribution} \\
  STOR-i Centre for Doctoral Training \\
  Lancaster University, UK\\
  \texttt{h.moss@lancaster.ac.uk} \\
  \And
    Ryan-Rhys Griffiths$^*$\\
  Department of Physics \\
  University of Cambridge, UK\\
  \texttt{rrg27@cam.ac.uk} \\
}

\begin{document}
\maketitle
\vspace{-0.6cm}
\begin{abstract}
We present FlowMO: an open-source Python library for molecular property prediction with Gaussian Processes. Built upon GPflow and RDKit, FlowMO enables the user to make predictions with well-calibrated uncertainty estimates, an output central to active learning and molecular design applications. Gaussian Processes are particularly attractive for modelling small molecular datasets, a characteristic of many real-world virtual screening campaigns where high-quality experimental data is scarce. Computational experiments across three small datasets demonstrate comparable predictive performance to deep learning methods but with superior uncertainty calibration.
\end{abstract}

\vspace{-0.6cm}
\section{Introduction}
\vspace{-0.3cm}
In the early stages of exploring a new class of drug molecules or molecular material, there is often only a small quantity of high-quality experimental data available \citep{thawani2020photoswitch, 2018_Griffiths}. In contrast to the big data regime of established molecule classes, where much of the information regarding performance has already been acquired, it is the small data regime which holds most promise for techniques such as Bayesian optimization \citep{2018_Bombarelli, 2020_Korovina, 2020_Griffiths, 2020_Tripp, Moss2020} and active learning \citep{2019_Zhang} to expedite the rate at which novel and performant molecules are discovered. 

The current trend in molecular machine learning research has been to leverage Bayesian deep learning to provide the uncertainty estimates which underpin active learning and Bayesian optimization methods \citep{2019_Ryu, 2019_Zhang, 2020_Hwang, 2020_Scalia}. Deep learning methods, however, are often not the model of choice for small datasets. In fact, leading experts on deep learning have expressed a preference for Gaussian Processes (GPs) \citep{rasmussen2004gaussian} for small datasets \citep{2011_Bengio}. Furthermore, Bayesian deep learning methods rely on approximate inference in order to produce uncertainty estimates. In contrast, GPs admit exact inference at the expense of computationally prohibitive cost for very large datasets. In this paper, we compare calibration of the uncertainty estimates of GPs against a popular Bayesian deep learning method as well as against the recently-introduced attentive neural process \citep{kim2019attentive}. 

Applying GPs to molecules in non-trivial, as the vast majority of existing applications of GPs assume inputs of low and fixed dimension. This assumption is not satisfied for the popular molecular representations of fingerprints \citep{rogers2010extended} and SMILES strings \citep{1988_Weininger}, sparse high-dimensional bit vectors and strings of variable length, respectively. To build GPs over molecules, FlowMO provides GPU-supported implementations of the Tanimoto and string kernels, providing a user-friendly way to make probabilistic predictions, whilst avoiding the expensive architecture tuning and model optimization often required to find effective deep learning models.

Our primary contributions can be summarized as follows. (1) We propose a GP framework for molecular property prediction using  Tanimoto and string kernels. (2) We provide an open-source Python implementation built upon GPflow\citep{2017_GPflow} and RDKit \citep{landrum2013rdkit}, addressing the absence of GP support in popular libraries such as DeepChem \citep{2019_Ramsundar}, DGL-LifeSci \citep{2020_Wang} and ASAP \cite{2020_Cheng}. (3) We compare FlowMO with established baselines across three benchmark datasets with plans for a more extensive comparison of uncertainty calibration.

\vspace{-0.3cm}
\section{Molecular Representations}
\vspace{-0.3cm}
\label{sec:representations}

\begin{figure}[t]
\centering
\begin{minipage}[t]{.46\textwidth}
  \centering
\includegraphics[width=\textwidth]{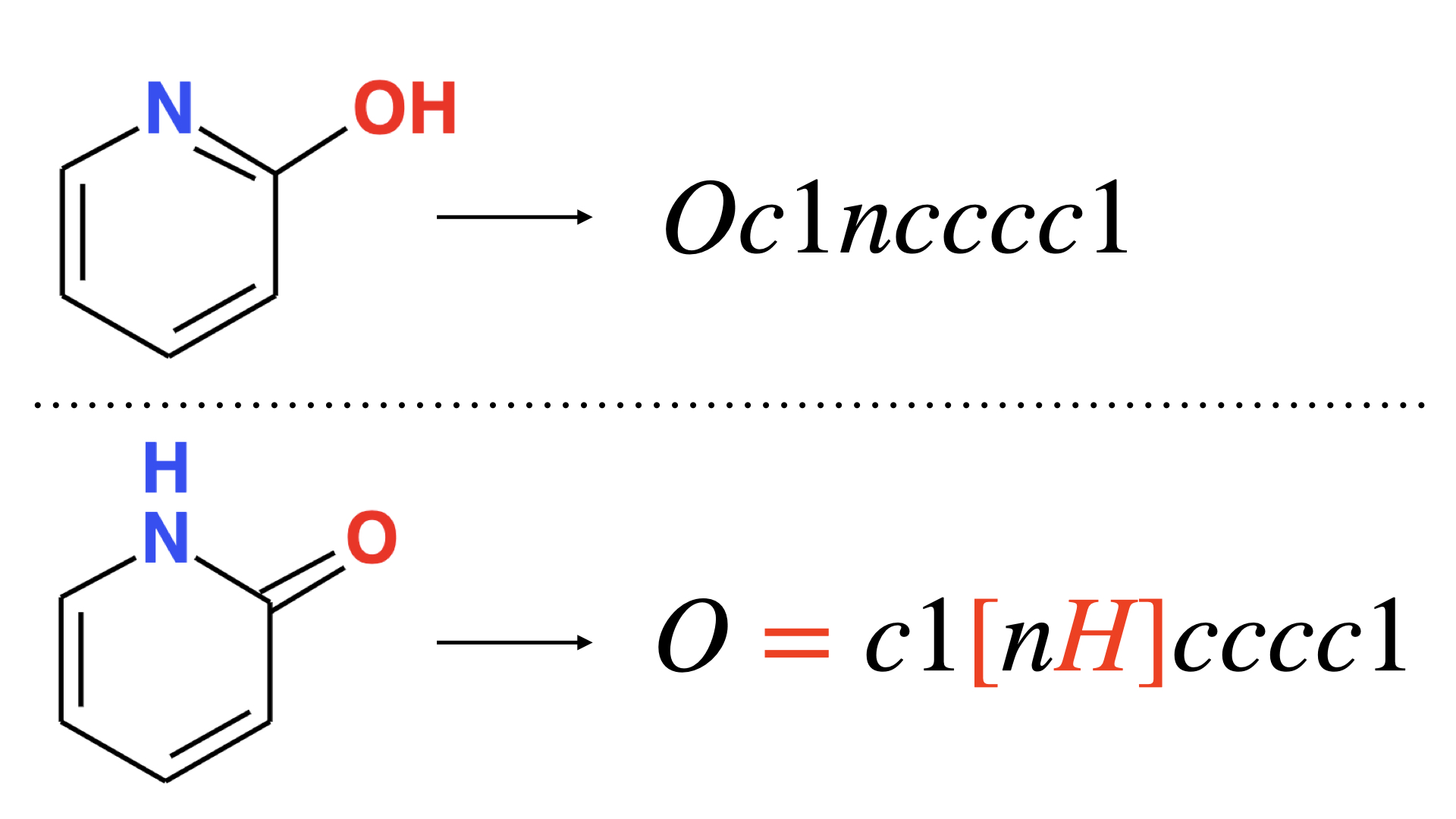}
\caption{SMILES strings for structurally similar molecules have local differences (red) but common non-contiguous sub-sequences (black).}
\label{fig:SMILES}
\end{minipage}%
\hfill
\begin{minipage}[t]{.53\textwidth}
  \centering
  \includegraphics[width=\textwidth]{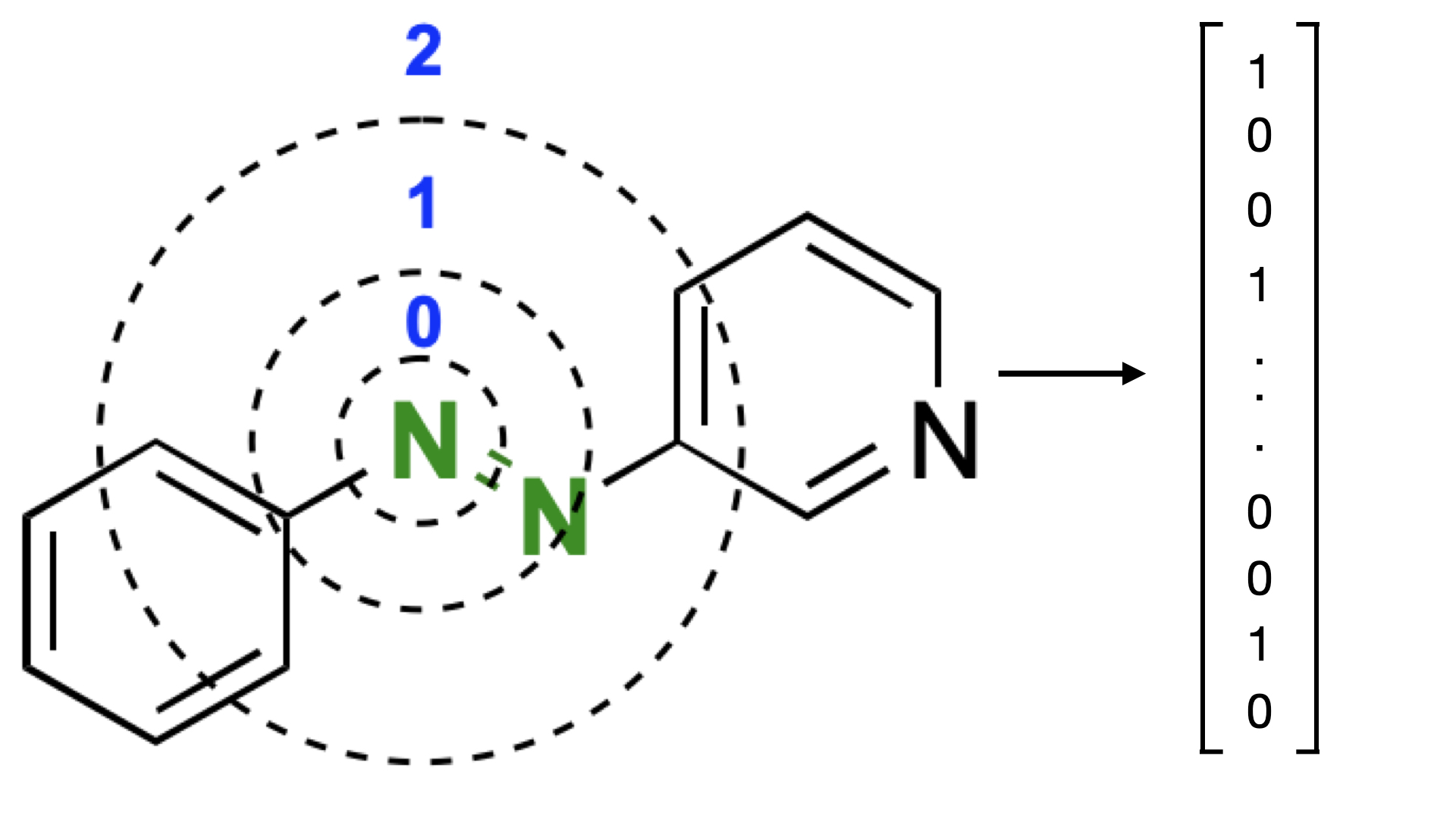}
  \captionof{figure}{Fingerprints provide high-dimensional binary representations of molecules by encoding fragments whose size is dictated by a pre-specified bond radius.}
    \label{fig:fingerprints}
\end{minipage}
\vskip -0.5cm
\end{figure}

To apply GPs to molecules, we need a meaningful way to represent molecules. FlowMO currently supports two popular representations: SMILES and ECFP fingerprints.

\textbf{SMILES strings}. The Simplified Molecular-Input Line-Entry System (SMILES) \citep{1988_Weininger} is a procedure for expressing molecules as character strings. Each SMILES string is assembled by traversing the molecule's  chemical graph. Two SMILES strings for structurally similar pyridine derivatives are presented in Figure \ref{fig:SMILES}. The alphabet of SMILES strings contains letters representing aliphatic and aromatic atoms $(B, C, N, O, S, P, F, Cl, Br, I, b, c, n, o, s, p)$, parentheses and integers, as well as  additional ASCII symbols representing bonds and the presence of rings $(-,=,\#,\&,/,\setminus,.,\%)$. 
 
\textbf{ECFP fingerprints}. ECFP fingerprints are high-dimensional bit vector representations of molecules \citep{christie1993structure} designed to search chemical databases and in the context of kernel-based classification have been used in conjunction with support vector machines \citep{ralaivola2005graph}. In this work, we consider extended-connectivity (ECFP) fingerprints \citep{rogers2010extended} as generated by the RDKit library \citep{landrum2013rdkit}. ECFP fingerprints are calculated with an iterative procedure. First, each atom is represented in terms of its local properties, for example its atomic mass and valency. These local representations are then updated based on the representations of its neighbors, iteratively building representations of all fragments containing as many atoms as the chosen bond radius of the fingerprint. Finally, duplicates structures are removed and the remaining representations are hashed into a unique bit vector (see Figure \ref{fig:fingerprints}). For our experiments, we use 2048-bit ECFP fingerprints with a bond radius of 3.

\vspace{-0.3cm}
\section{Molecular Property Prediction with Gaussian Processes}
\vspace{-0.3cm}
Given a training set of (possibly noisy) experimentally-determined molecular properties and a kernel function $k$ measuring intermolecular similarity, Gaussian processes (See \cite{rasmussen2004gaussian} for a comprehensive introduction) provide tractable Gaussian predictive distributions for the properties of any out-of-sample target molecule. In FlowMO, we provide a kernel for each of our supported molecular representations (as introduced above):  a Sub-sequence String Kernel (SSK) and a Tanimto Kernel (TK) suitable for measuring the similarity between SMILES strings of varying length and  high-dimensional binary fingerprint vectors, respectively. 

The computational complexity of training and inference for the GP are $O(n^3+n^2C)$ and $O(nC)$, where $n$ is the number of training molecules and $C$ is the cost of a single kernel evaluation. Although there exist many methods that reduce the computational complexity of GPs (for example \cite{hensman2013gaussian}),  they are designed with smooth input spaces in mind and do not yet support molecular data. In practice, we found standard GPs to be adequate for the datasets considered in this paper, and furthermore they can be comfortably applied to larger data sets using FlowMO's GPU support.

% , or that we have exact evaluations (i.e $\sigma^2_o=0$),  our GP conditioned on the training data $D$ produces a tractable predictive distribution for the target property of any possible molecule $\textbf{m}^*$.  $\mu(\textbf{m}^*)= \textbf{k}(\textbf{m}^*)^T(\textbf{K}+\sigma^2_oI)^{-1}\textbf{y}$ and variance $\sigma^2(\textbf{m}^*)=k(\textbf{m}^*,\textbf{m}^*)-\textbf{k}(\textbf{m}^*)^T(\textbf{K}+\sigma_o^2I)^{-1}\textbf{k}(\textbf{m}^*)$,
% where we have defined 
% $\textbf{K}=\left[k(\textbf{m},\textbf{m}')\right]_{(\textbf{m},\textbf{m}')\in D}$, $\textbf{k}(\textbf{m}^*)=\left[k(\textbf{m},\textbf{m}^*)\right]_{ \textbf{m}\in D}$ and $\textbf{y}=[y_i]_{i=1,..,n}$. 

\textbf{String Kernels}. Sub-sequence String Kernels \citep{lodhi2002text,cancedda2003word} measure the similarity of strings through the number of shared sub-strings, naturally supporting variable length inputs. The considered sub-strings can be non-contiguous, yielding a rich contextual model of string data that encapsulates the non-contiguity known to be important when characterising inter-SMILES similarity (Figure \ref{fig:SMILES}). To avoid enumeration of an exponentially-scaled feature space, SSKs exploit the kernel trick through an efficient dynamic programming algorithm, allowing individual kernel calculations between two strings $\textbf{s}$ and $\textbf{s}'$ to be calculated in $C=O(\max(|\textbf{s}|,|\textbf{s}'|)^2)$ operations. FlowMO's SSK implements the vectorized formulations of  \cite{beck2017learning} and \cite{Moss2020}, enabling efficient computation on GPUs. FlowMO's SSKs have two kernel parameters $\lambda_m$ and $\lambda_g$ controlling the relative weighting of long and/or highly non-contiguous sub-strings. In all our experiments, we set $n=5$ and learn appropriate values for $\lambda_g$ and $\lambda_m$ through the standard practice of maximising the marginal likelihood of the GP.

% An $n^{th}$ order SSK measures the similarity of two molecules $\textbf{m}$ and $\textbf{m}'$ via their SMILES strings $\textbf{s}$ and $\textbf{s}'$ as
% \begin{align}
% {k}^{SSK}_n(\textbf{m},\textbf{m}')=\sigma^2 \frac{ \textbf{c}(\textbf{s})\cdot\textbf{c}(\textbf{s}')}{||\textbf{c}(\textbf{s})||||\textbf{c}(\textbf{s}')||}\; \mbox{ where} \; \textbf{c}(\textbf{s})=\left[\lambda_m^{|\textbf{u}|}\sum_{1<i_1<..<i_{|\textbf{u}|}<|\textbf{s}|}\lambda_g^{i_{|\textbf{u}|}-i_1} \mathds{1}_{\textbf{u}}((s_{i_1},..,s_{i_{|\textbf{u}|}}))\right]_{\textbf{u}\in\Sigma^n},
% \label{eq:SSK}\nonumber
% \end{align}
% where $\Sigma^n$ denotes the set of all possible ordered collections containing up to $n$ characters from the SMILES alphabet, $\mathds{1}_{\textbf{x}}(\textbf{y})$ is the indicator function checking if the strings $\textbf{x}$ and $\textbf{y}$ match, and $\sigma^2\in\mathds{R}^+$, $\lambda_m\in[0,1]$ and $\lambda_g\in[0,1]$ are kernel hyper-parameters. $c(\textbf{s})_i$ measures the contribution of each sub-sequence $\Sigma^n_i$ to string $\textbf{s}$, 

\textbf{Tanimoto Kernels}. To build a GP model over fingerprint representations of molecules, we employ a kernel based on the Tanimoto similarity measure from the chemoinformatics literature \citep{gower1971general, fligner2002modification, ralaivola2005graph}. Note that ECFP fingerprints contain $\gg1000$ binary entries, each flagging the occurrence of particular features, and so have length and sparsity unsuitable for standard GP kernels designed for low-dimensional continuous spaces. The Tanimoto Kernel (TK) measures the similarity between two molecules via their fingerprints by the number of features present in both molecules normalized by the number of features occurring separately, i.e $ k(\textbf{f},\textbf{f}')=\sigma^2 \textbf{f}\cdot\textbf{f}'/ (\textbf{f}\cdot\textbf{f}  + \textbf{f}'\cdot\textbf{f}'  -  \textbf{f}\cdot\textbf{f}') $, where $\textbf{f}$ denotes a fingerprint representation and $\sigma^2$ is a single kernel hyper-parameter. The complexity of each kernel evaluation is simply $C=O(|\textbf{f}|)$.

% \begin{align}
%     k^{Tanimoto}(\textbf{m},\textbf{m}')=\sigma^2\frac{ \textbf{f}\cdot\textbf{f}' }{ \textbf{f}\cdot\textbf{f}  + \textbf{f}'\cdot\textbf{f}'  -  \textbf{f}\cdot\textbf{f}_' },
%     \label{eq:tanimoto}\nonumber
% \end{align}
% where $\sigma^2$ is a single kernel hyper-parameter. As the entries of print vectors are binary flags denoting feature occurrences, the TK can be interpreted as measuring similarity 

\vspace{-0.3cm}
\section{Experimental Validation}
\vspace{-0.3cm}
We  now  evaluate FlowMO (available at \href{https://anonymous.4open.science/r/f160a0a2-0161-4d31-ba55-2e3aab2133d3/}{anonymized link}) across three small molecular property prediction datasets.
\begin{enumerate}
\itemsep0em 
    \item The Photoswitch Dataset \citep{thawani2020photoswitch}: a collection of 392 photoswitch molecules and their experimental \textit{E} isomer $\pi=\pi^*$ transition wavelengths.
    \item ESOL \citep{delaney2004esol}: the logarithmic aqueous solubility for 1128 organic small molecules.
    \item FreeSolv \citep{mobley2014freesolv}: experimental hydration free energies for  642 molecules.
\end{enumerate}
Both ESOL and FreeSolv are part of the MoleculeNet dataset collection widely used for the benchmarking of molecular property prediction models \citep{wu2018moleculenet}, whereas the Photoswitch Dataset is a recent benchmark designed to specifically investigate the properties of light-activated molecules.

We compare the performance of our SSK and TK GPs with a range of existing molecular property prediction models. We report the scores of the best found model during the extensive testing of molecular graph-based models detailed in \cite{wu2018moleculenet} namely a graph convolutional network (GCN) \citep{kipf2016semi} for the Photoswitch Dataset and a message passing neural network (MPNN) \citep{gilmer2017neural} for ESOL and FreeSolv. These scores are regarded as very strong benchmarks, only recently improved upon for ESOL and FreeSolv by SMILES-X \citep{lambard2020smiles}. SMILES-X is a sophisticated model that relies on extensive hyperparameter optimization and requires data augmentation via augmented random SMILES. We include the performance of SMILES-X pre and post augmentation. Across all three data sets, we also consider a black-box alpha divergence minimization Bayesian Neural Network (BNN) \citep{2016_Lobato} and an Attentive Neural Process (ANP) \citep{kim2019attentive} applied to fingerprint representations. Our ANP and BNN follow the exact implementation described by \citep{thawani2020photoswitch}, with hyperparameters tuned by grid search. Further details are available in the supplementary information. Reported scores for all methods are based on 20 random train/validation/test splits in an 80:10:10 ratio, except for the GPs which need no validation set and so can be applied to 90:10 train/test splits.

% Node features include
% one-hot representations of atom-type, atom degree, the number of implicit hydrogen atoms attached
% 4
% to each atom, the total number of hydrogen atoms per atom, atom hybridization, the formal charge
% and number of radical electrons on the atom. Edge features contain one-hot encodings of bond-type
% and Booleans indicating the stereogenic configuration of the bond and whether the bond is conjugated
% or in a ring. For the GCN we use two hidden layers with 32 hidden units and ReLU activations,
% applying BatchNorm [68] to both layers. The remaining parameters are the default library values.

\vspace{-0.2cm}
\subsection{Predictive Performance}
\vspace{-0.2cm}
Table \ref{tables} demonstrates the predictive performance of the models, showing the mean and a single standard deviation of the root-mean-squared error (RMSE). The best GP models outperform the ANP and BNN, and achieve comparable performance to both the MoleculeNet collection and the pre-augmentation SMILES-X models. Only the substantially more computationally demanding augmented SMILES-X model is able to significantly outperform the GPs and only on the FreeSolv data. The SSK GP outperforms the TK GP on all but the Photoswitch Dataset, suggesting that SMILES strings are useful representations for predicting aqueous solubility and hydration free energies, but ECFP fingerprints are especially informative for modeling transition wavelengths highlighting the importance of feature engineering for molecular property prediction tasks.

\begin{table}[t]
\caption{RMSE of the models across the three datasets, with the scores of the best GP model and best overall model highlighted .\label{tables}}
\centering
% \begin{tabular}{@{}l|llll@{}}
% \toprule
%  & Photoswitch & ESOL & FreeSolv \\ \midrule
% SSK GP (SMILES)&  26.0 \pm 0.8 & \textbf{0.67} \pm \textbf{0.01} & \textbf{1.39} \pm 0\textbf{.06} \\
% TK GP (Fingerprints)  & \textbf{22.6 }\pm \textbf{0.9 }  & 1.01 \pm 0.02 & 1.93 \pm 0.04 \\\bottomrule
% % #RBF GP (Fragments)  &    & 0.90 \pm 0.01 & 1.47 \pm 0.03 \\
% ANP & ?  & ? &  ?\\
% BNN & ?  & ? &  ?\\
% GCN & ?  & ? &  ?\\ 
% SMILES-X & -  & 0.70 \pm 0.05&  1.14 \pm 0.17\\ 
% SMILES-X (Augm) & -  & \textbf{0.57} \pm \textbf{0.07} &  \textbf{0.81} \pm \textbf{0.22}\\ 
% MoleculeNet& -  & \textbf{0.58} \pm \textbf{0.03} &  1.15 \pm 0.02\\
% \end{tabular}
\begin{tabular}{@{}l|llll@{}}
\toprule
 & Photoswitch & ESOL & FreeSolv \\ \midrule
SSK GP (SMILES)&  26.0 $\pm$ 3.6 & \textbf{0.65} $\pm$ 0.04 & \textbf{1.29} $\pm$ 0.22\\
TK GP (Fingerprints)  & \textbf{22.6 }$\pm$ 4.0  & 0.98 $\pm$ 0.08 & 1.85 $\pm$ 0.10 \\\bottomrule
ANP & 27.2 $\pm$  3.7 & 1.32 $\pm$ 0.13 &  2.65 $\pm$ 0.47\\
BNN & 25.5 $\pm$ 5.0 & 1.01 $\pm${0.11} &  1.92 $\pm$ 0.20\\
MoleculeNet& \textbf{22.0}  $\pm$ 3.5  & \textbf{0.58} $\pm$ 0.03 &  1.15 $\pm$ 0.02\\
SMILES-X & -  & 0.70 $\pm$ 0.05&  1.14 $\pm$ 0.17\\ 
SMILES-X (Augm) & -  & \textbf{0.57} $\pm$ 0.07 &  \textbf{0.81} $\pm$ 0.22\\ 

\end{tabular}
\end{table}
\vspace{-0.2cm}
% \todo{We should comment why missing entries in table?}
\subsection{Calibration}
\vspace{-0.2cm}

\begin{figure}[t]
\setlength{\tabcolsep}{1pt}
\begin{tabular}{ccc}
  \includegraphics[height=32mm]{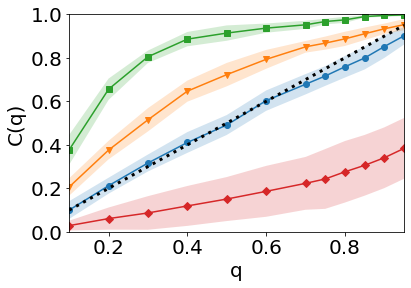} &   \includegraphics[height=32mm]{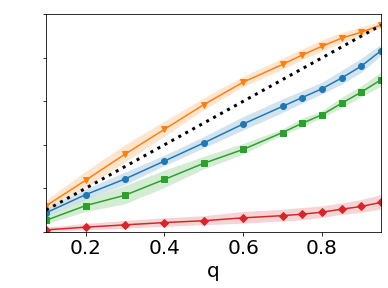} & \includegraphics[height=32mm]{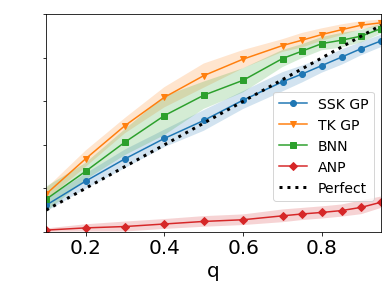} 
  \vspace{-10pt}\\
Photoswitch &  ESOL & FreeSolv
\end{tabular}
\caption{Calibration scores for the probabilistic models. We see reliable calibration among our GP models, with the SSK GP achieving almost perfect calibration (as represented by the diagonal black dotted line) for Photoswitch and FreeSolv.}
\label{calib_fig}
\vspace{-0.6cm}
\end{figure}

To analyze the calibration achieved by the predictive distributions provided by the probabilistic models (only the GPs, BNN and ANP), we define a calibration score function \begin{align}
    \textrm{C(q)} = \frac{1}{|\mathcal{T}|}\sum_{m\in\mathcal{T}}\left[\mathds{1}\left(\left\vert\frac{\hat{y}(m)-y(m)}{\hat{\sigma}(m)}\right\vert<\Phi^{-1}\left(\frac{1+q}{2}\right)\right)\right] \nonumber
\end{align}
based on cross-validatory predictive p-values \citep{marshall2003approximate, leslie2007general}. $y(m)$, $\hat{y}(m)$ and $\hat{\sigma}(m)$ represent true values, predictive means and predictive standard deviations for each test molecule $m\in\mathcal{T}$, and $\Phi^{-1}$ is the inverse of the standard Gaussian cumulative distribution function. The indicator $\mathds{1}$ is activated only when the true value is contained in the model's $q*100\%$ confidence interval. Therefore, perfect calibration at the $q^{th}$ quantile corresponds to $C(q)=q$. $C(q)>q$ indicates under-confidence through overly large uncertainty estimates (limiting the strength of conclusions that can be drawn from the model) whereas $C(q)<q$ denotes over-confidence (leading to reckless decisions downstream). We plot $C(q)$ for our probabilistic models as Figure \ref{calib_fig}.

\vspace{-0.4cm}

\section{Future Work}
\vspace{-0.3cm}
We have presented an analysis of GPs for molecular property prediction and uncertainty quantification. Future work will extend this analysis in three ways. Firstly, as string kernels operate directly on SMILES strings, we will explore if the data augmentation strategy exploited by SMILES-X can yield a similar performance boost for our SSK GPs.  Secondly, we wish to consider GPs built upon additional kernels belonging to the broad class of convolutions kernels that includes our SSK. For example, graph kernels \citep{vishwanathan2010graph} could be used to define GPs directly over molecular graphs, enabling a comparison with recent work on graph-based deep learning methods \citep{2019_Yang,2020_Hwang}. Finally, FlowMO will allow us to apply the extended Bayesian optimization methods designed for GP models to further improve efficiency in automatic wet-lab workflows
\citep{macleod2020self}. Potential extensions include batch \citep{gonzalez2016batch}, multi-task \citep{swersky2013multi}, multi-fidelity \citep{moss2020mumbo} and  multi-objective \citep{hernandez2016predictive} optimization, as well as optimization with controllable experimental noise \citep{moss2020bosh}.

\def\bibfont{\small}
\bibliography{references}

\appendix

\section{Broader Impact}

As mentioned in the main body of the paper, calibrated uncertainty estimates are highly important for the successful implementation of techniques which depend on them such as Bayesian Optimization and active learning. To this end, Gaussian Processes are an important, yet somewhat neglected method (in the molecular machine learning domain) for producing well-calibrated uncertainty estimates for small datasets. It is our hope that through the provision of a bespoke library for Gaussian Processes for molecular property prediction, we can enhance the efficacy of downstream Bayesian Optimization and active learning campaigns in discovering promising drug molecules and molecular materials in the laboratory.

\section{Hyper-parameter Settings}

For the ESOL and FreeSolv data, hyper-parameters were selected by validation on an 80/10/10 train/validation/test split. For the Photoswitch data a split of 80/20 train/test was required in order to maintain consistency with the results reported in \cite{thawani2020photoswitch}. In this latter case, validation was performed by taking the original train set (80) and splitting this in the ratio of 90/10. Following hyper-parameter selection, the models were then re-trained using these hyper-parameters on the original train set. 

For the BNN, a grid search was performed over the number of nodes per layer, the learning rate, the batch size as well as the number of iterations for the ADAM optimizer. The $\alpha$ hyper-parameter was fixed at 0.5 and the number of samples was fixed at 100. ReLU activations were the only activation functions considered and the number of layers was fixed to 2. 

For the ANP, a grid search was performed over the number of nodes in the deterministic encoder, the number of nodes in the latent encoder and the number of nodes in the decoder in addition to the learning rate, the batch size and the number of training iterations. The dimensionality of the context encoding was fixed to 8. The number of layers for the deterministic encoder, the latent encoder and the decoder was fixed to 2 in each case. The number of samples to take of the context set was fixed to 10.

\end{document}